\documentclass[11pt,table,dvipsnames]{article}

\usepackage[preprint]{neurips_2024}
\usepackage[utf8]{inputenc}
\usepackage[T1]{fontenc}
\usepackage{hyperref}
\usepackage{url}
\usepackage{booktabs}
\usepackage{amsfonts}
\usepackage{nicefrac}
\usepackage{microtype}
\usepackage{xcolor}
\usepackage{times}
\usepackage{latexsym}
\usepackage{CJKutf8}
\usepackage{graphicx}
\usepackage{bigstrut}
\usepackage{multirow}
\usepackage{amsmath}
\usepackage[normalem]{ulem}
\usepackage{multicol}
\usepackage{tipa}
\usepackage{bm}
\usepackage[ruled,linesnumbered,vlined]{algorithm2e}
\usepackage{paralist}
\usepackage{arydshln}
\usepackage{verbatim}
\usepackage{csquotes}
\usepackage{xspace}
\usepackage{mdwlist}
\usepackage{subfigure}
\usepackage{makecell}
\usepackage{array}
\usepackage{enumitem}

\usepackage{colortbl}
\usepackage{dsfont}
\usepackage{tabularx}
\usepackage{bbm}
\usepackage{tikz}
\usepackage{pgfplots}
\usepackage{wrapfig}

\usepackage{listings}
\usepackage{amsmath}%
\usepackage{MnSymbol}%
\usepackage{wasysym}%
\usepackage{algpseudocode}
\usepackage{natbib}
\setcitestyle{numbers,square}
\usepackage[nameinlink,capitalise]{cleveref}
\hypersetup{
    colorlinks,
    linkcolor={blue!80!black},
    citecolor={blue!80!black},
}
\newcolumntype{C}[1]{>{\centering\arraybackslash}p{#1}}
\newcolumntype{L}[1]{>{\arraybackslash}p{#1}}

\newcommand{\method}{\textsc{SelfGoal}\xspace}
\newcommand{\tree}{\textsc{GoalTree}\xspace}

\newcommand{\ie}{\textit{i.e.}\xspace}
\newcommand{\eg}{\textit{e.g.}\xspace}

\newcommand{\goal}[1]{``\texttt{#1}''}

\newcommand{\cjj}[1]{\textcolor{blue}{{[cjj: #1]}}}

\definecolor{customcolor}{rgb}{0.2, 0.4196, 0.8039}

\makeatletter
\def\adl@drawiv#1#2#3{%
        \hskip.5\tabcolsep
        \xleaders#3{#2.5\@tempdimb #1{1}#2.5\@tempdimb}%
                #2\z@ plus1fil minus1fil\relax
        \hskip.5\tabcolsep}
\newcommand{\cdashlinelr}[1]{%
  \noalign{\vskip\aboverulesep
           \global\let\@dashdrawstore\adl@draw
           \global\let\adl@draw\adl@drawiv}
  \cdashline{#1}
  \noalign{\global\let\adl@draw\@dashdrawstore
           \vskip\belowrulesep}}
\makeatother

\title{\method: Your Language Agents Already Know How to Achieve High-level Goals}

\author{
 Ruihan Yang$^\heartsuit$, 
 Jiangjie Chen$^\heartsuit$\thanks{Corresponding authors.}, 
 Yikai Zhang$^\heartsuit$, 
 Siyu Yuan$^\heartsuit$, 
 Aili Chen$^\heartsuit$, \\
 \bf Kyle Richardson$^\clubsuit$
 Yanghua Xiao$^\heartsuit$,
 Deqing Yang$^\heartsuit$\footnotemark[1]\\
 $^\heartsuit$Fudan University \quad $^\clubsuit$Allen Institute for AI\\
    \small \texttt{\{rhyang17, jjchen19, alchen20, shawyh, deqingyang\}@fudan.edu.cn} \\
    \small \texttt{\{ykzhang22, syyuan21\}@m.fudan.edu.cn}\quad \texttt{kyler@allenai.org}
}

\begin{document}

\maketitle

\begin{abstract}
Language agents powered by large language models (LLMs) are increasingly valuable as decision-making tools in domains such as gaming and programming. 
However, these agents often face challenges in achieving high-level goals without detailed instructions and in adapting to environments where feedback is delayed. 
In this paper, we present \method, a novel automatic approach designed to enhance agents' capabilities to achieve high-level goals with limited human prior and environmental feedback. 
The core concept of \method involves adaptively breaking down a high-level goal into a tree structure of more practical subgoals during the interaction with environments while identifying the most useful subgoals and progressively updating this structure. 
Experimental results demonstrate that \method significantly enhances the performance of language agents across various tasks, including competitive, cooperative, and deferred feedback environments.\footnote{Project page: \url{https://selfgoal-agent.github.io}.}

\end{abstract}

\section{Introduction}
\label{sec:Intro}

The advancement of large language models (LLMs)~\citep{brown2020language,openai2022chatgpt,openai2024gpt4} has enabled the construction of autonomous \textit{language agents} (or LLM-based agents) to solve complex tasks in dynamic environments without task-specific training.
In reality, these autonomous agents are often tasked with very broad, high-level goals, such as \goal{winning the most money} or \goal{succeeding in a competition}, whose ambiguous nature and delayed reward raise great challenges for autonomous task-solving.
More importantly, it is not practical to frequently train these models to adapt to new goals and tasks~\citep{zheng2023agents, khot2023decomposed, prasad2024adapt}.
Therefore, a critical question arises: 
\textit{How can we enable autonomous language agents to consistently achieve high-level goals without training?}

Previous works focus on creating two types of auxiliary guidance in the instructions for language agents to achieve high-level goals in tasks: prior task decomposition and post-hoc experience summarization.
The former involves decomposing the task before acting, utilizing prior knowledge from LLMs to break down high-level goals into more tangible subgoals related to specific actions at hand~\cite{yuan-etal-2023-distilling,zheng2023agents,singh2024twostep,liu2024delta}. 
However, this line of work does not ground these subgoals into the environment during interaction, resulting in the loss of empirical guidance. 
In contrast, the latter allows agents to interact directly with environments and summarize valuable experiences from history~\cite{madaan2023selfrefine,majumder2023clin,zhao2024expel, paul2024refiner}, \eg, ``X contributes to Y''.
However, the difficulty of inducing rules from experience causes the guidance to be simple and unstructured, making it difficult to prioritize or adjust strategies effectively.

A natural solution to combine the best of both worlds is to dynamically decompose the task and its high-level goal during interaction with the environment. 
This approach requires an agent to build and use guidelines that vary in detail and aspect. 
A tree structure is ideal for this requirement, as it allows hierarchical organization, providing both broad overviews and detailed guidance as needed.
However, this approach presents two major challenges:
\begin{inparaenum}[\it 1)]    
    \item Not all nodes are relevant to the current context during task execution, which requires selecting the most suited nodes to guide current actions.
    For example, \goal{watch for bargains} is a more prudent choice than \goal{bid on the most expensive item} when budget is tight;
    \item The granularity of guidance provided by nodes increases with tree depth, yet the appropriate detail level varies across scenarios, making a fixed tree depth not general. 
    For example, a generic guideline like \goal{earn more money} is not useful in auctions.
\end{inparaenum}

\begin{figure*}[t]
    \centering
    \includegraphics[width=0.95\textwidth]{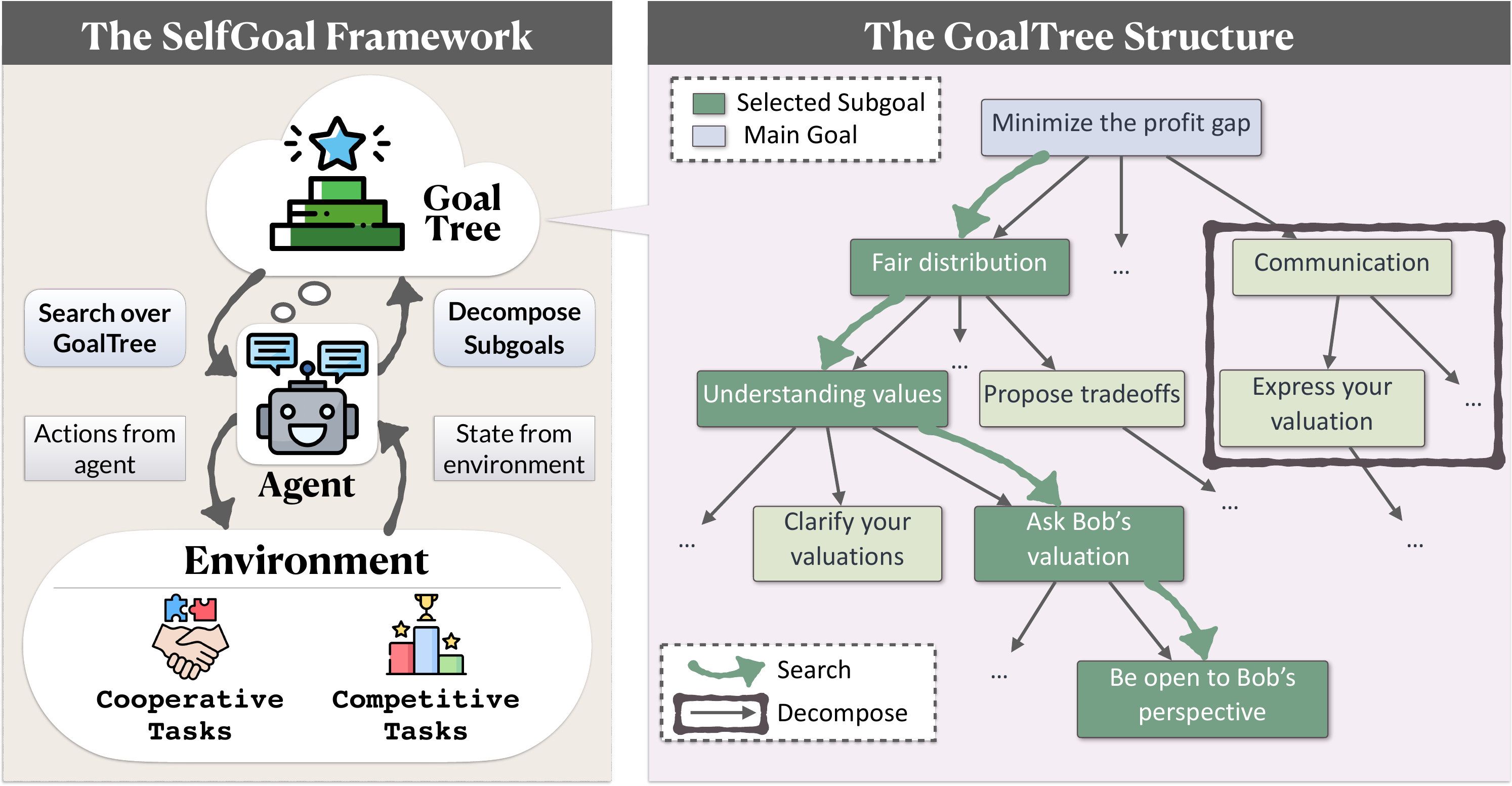}
    \caption{An overview of \method, illustrated with a bargaining example. The agent interacts with environments, and make actions based on environmental feedback and the \tree dynamically constructs, utilizes and updates with Search and Decompose Modules.}
    \label{fig:workflow}
    \vspace{-8pt}
\end{figure*}

To tackle these challenges, we propose \method, a self-adaptive framework for a language agent to utilize both prior knowledge and environmental feedback to achieve high-level goals.
The main idea is to build a tree of textual subgoals, where agents choose appropriate ones as the guidelines to the prompt based on the situation.
Specifically, as shown in Figure~\ref{fig:workflow}, \method is featured with two main modules to operate a \tree, which is constructed, updated, and utilized during task execution:
\begin{inparaenum}[\it 1)]
    \item \textbf{Search Module} is prompted to select the top-K most suited nodes of goals based on the provided current state and existing nodes in \tree, which utilizes the prior knowledge of LLMs;
    \item \textbf{Decomposition Module} breaks down a goal node into a list of more concrete subgoals as subsequent leaves, ensuring an adaptive self-growth of \tree. Note that we filter out the redundant nodes during decomposition based on the textual similarity between new ones and the existing nodes of goals;
    \item \textbf{Act Module} takes as input the selected subgoals as guidelines, and prompts LLMs for actions for the current state.
\end{inparaenum}
Extensive experiments in various competition and collaboration scenarios show that \method provides precise guidance for high-level goals and adapts to diverse environments, significantly improving language agent performance. 

In summary, our contributions in this paper are as follows:
\begin{itemize}[leftmargin=*]
    \item We target the challenge of enabling autonomous language agents to consistently achieve high-level goals without the need for frequent retraining.
    \item We introduce \method, a self-adaptive framework that constructs, updates, and utilizes a \tree to dynamically decompose a task's high-level goals into subgoals during interaction with the environment. 
    \item We conduct extensive experiments in both collaborative and competitive scenarios where agents tend to deviate from their goals. The results demonstrate that \method significantly enhances the capability of language agents to adhere to high-level goals consistently.
\end{itemize}

\section{Related Work}
\label{sec:Related}

\paragraph{Learning from Feedback}
LLMs have become a promising tool for building goal-directed language agents~\citep{huang2022language}. 
With textual input that includes the world state, task, and interaction history, language agents are to decide the next action to achieve a goal~\citep{lin2023swiftsage, yao2023react}. 
Studies have explored enhancing the reasoning and planning abilities of language agents through feedback from environments. 
For example, Reflexion~\citep{shinn2023reflexion} enables an agent to reflect on its failures and devise a new plan that accounts for previous mistakes. 
Similarly, Voyager~\citep{Wang2023VoyagerAO} operates in Minecraft, developing a code-based skill library from detailed feedback on its failures. 
Recent works~\citep{majumder2023clin,nottingham2024skill} analyze both failed and successful attempts, summarizing a memory of causal abstractions.
However, learnings directly from feedback are often too general and not systematic, making it difficult to prioritize strategies effectively.

\paragraph{LLMs for Decision Making}
LLMs are increasingly used as policy models for decision-making in interactive environments such as robotics~\citep{ahn2022i, huang2022inner, liu2023llmp}, textual games~\citep{wang-etal-2023-plan,zhang2024timearena,xie2024travelplanner,ma2024red}, and social tasks~\citep{zhou2024sotopia}. 
However, the goals in these environments, like \goal{find a fruit} in ScienceWorld~\citep{wang2022scienceworld}, are often simple and specific. 
For long-term, high-level goals, LLMs struggle to perform effectively \citep{hoang2021successor,huang2019mapping}, and additional modules are needed for support\citep{zheng2023agents}. 
In our work, we use a method that does not require updating LLM parameters, enabling language agents to consistently pursue high-level goals during interactions with environments.

\paragraph{Decomposition and Modularity} 
Decomposing complex decision-making tasks into sub-tasks is a traditional method that enhances LLM task-solving capabilities \citep{barto2003recent, pellier2023hddl}. 
Approaches like Hierarchical Task Networks leverage domain knowledge, including a hand-specified library of plans, to simplify complex problems \citep{erol1994htn}. 
Recently, some studies have assigned LLMs the role of decomposing goals. 
For example, Decomposed Prompting \citep{khot2022decomposed} uses a few-shot prompting approach to tackle multi-step reasoning tasks by breaking them into a shared library of prompts. 
OKR-Agent \citep{zheng2023agents} utilizes self-collaboration and self-correction mechanisms, supported by hierarchical agents, to manage task complexities. 
ADAPT \cite{prasad2024adapt} enables LLMs to recursively re-decompose goals based on feedback in decision-making tasks. 
However, these approaches often decompose tasks before interaction with the environments, resulting in a lack of grounded, dynamic adjustment.
To address this, we aim to combine modular goal decomposition with learning from environmental feedback.

\section{Methodology}
\label{sec:Method}

\begin{wrapfigure}{r}{0.46\textwidth}
\vspace{-20pt}
\hspace{10pt}
    \begin{minipage}{\linewidth}
    \centering
    \input{tables/algorithm}
    \end{minipage}
\vspace{-8pt}
\end{wrapfigure}

When executing complex tasks with high-level goals (\eg, \goal{forecast future stock prices}), humans usually decompose it into specific detailed subgoals (\eg, \goal{gather historical price data and adjust predictions based on recent market events}) for effective execution~\citep{Valerie2011}.
Inspired from this idea, we propose \method in this paper, which is a non-parametric learning approach for language agents to exploit and achieve high-level goals.
\method conducts a top-down hierarchical decomposition of the high-level goal, with a tree of nodes representing useful guidance for decision-making. 

In this section, we first outline how \method works in $\mathsection$\ref{sec:overview}. 
Next, we explain the details of three key modules (Search, Decompose and Act) in \method that help maintain a tree of subgoals (\tree) in $\mathsection$\ref{sec:example} and guide task execution.

\subsection{Overview of \method}
\label{sec:overview}

\paragraph{Problem Formulation: Tasks with High-level Goals}
First, we formulate the features of our studied tasks, requiring an agent to interact with a dynamic environment and evaluated based on the achievement of the high-level goal.
We focus on the scenarios where an actor model $M_a$ aims to achieve a high-level goal $g_0$ in an environment $E$ through interaction. 
The policy employed by $M_a$ is denoted as $\pi_\theta$.
At each timestep $t$, $\pi_\theta$ generates an action $a_t$, and the environment $E$ returns a state $s_t$. 
This action-state pair $\{a_t, s_t\}$ is then utilized to update $\pi_\theta$.
Note that \method also supports accomplishing long-horizon tasks that do not always have immediate rewards.
In this case, only by completing the task $M_a$ will be evaluated with a score according to the achievement of the goal $g_0$.

\paragraph{Workflow of \method}
\method is a non-parametric learning algorithm for language agents, i.e., without parameter update.
The workflow of \method is shown at Algorithm~\ref{alg:adaptive_objective_tree_development}, which models $\pi_\theta = p$ by setting $p$ as the instruction prompt provided to $M_a$ (an LLM), i.e., $a_t = \texttt{LLM}(p_{t}, s_{t-1})$.
The policy $\pi_\theta$ is updated through the modifications to $p$, which is the modification of subgoal instructions $g_{i,j}$ ($i$-th layer, $j$-th leaf node) that best suit the current situation.
Concretely, \method is featured with two key modules, \textbf{Decomposition} and \textbf{Search} which construct and utilize a subgoal tree $\mathbb{T}$ respectively, namely \tree, to interact with the environment.
Setting the high-level goal of the task as the root node in \tree, 
\textbf{Search Module} finds the nodes that are helpful for the status quo, and \textbf{Decomposition Module} decomposes the node into subgoals as leaf nodes if it is not clear enough.

\subsection{Details in \method}
\label{sec:example}
\paragraph{Search: Identifying Useful Subgoals for Current Situation}
In the \textbf{Search Module} of \method, we ask the backbone LLM of the agent to identify the most appropriate subgoal for the current situation, e.g., ``Select $K$ most useful sub-goals that will help you reach your main goal in the current situation ...'' (see Appendix~\ref{app:selfgoal_prompt} for the complete prompt).
We represent the the current state $s_t$ for timestep $t$ as a description of the dialogue history of the interaction with the environment.
We also find the leaf nodes of each branch in \tree as the candidate subgoal list for LLMs to decide which ones that are useful.
The LLM then selects $K$ most suitable subgoals, followed by decomposition and the update of the instruction prompt $p_{t}$ at this step.

\paragraph{Decompose: Refine \tree to Adapt to the Environment}
Based on the current action-state pair $\{a_t, s_t\}$, \tree is updated through decomposition if it is not specific enough for useful guidance to the agent.
We use the backbone LLM to break down the selected subgoal $g_{i,j}$ in the \textbf{Search Module} (initially set to $g_0$). 
We prompt the LLM with the instruction such as ``What subgoals can you derive from $\{g_{i,j}\}$, based on $\{a_t, s_t\}$'', which generates a new set of subgoals $G$ (see also Appendix~\ref{app:selfgoal_prompt}). 
To control the granularity of these subgoals, we apply a \textit{filtering mechanism} that if the cosine similarity~\citep{rahutomo2012semantic} between a new subgoal and existing subgoals exceeds $\xi$, the current node will not be updated.
Otherwise, we add the new subgoals under the current node, thus expanding the \tree. 
Moreover, a \textit{stopping mechanism} is designed that if no new nodes are added to the \tree for $N$ consecutive rounds, the update is stopped.

\paragraph{Act: Utilizing Subgoals to Take Actions}
After getting the subgoals from \tree that are found by \method as useful, the agent updates the instruction prompt $p_t$ for the LLM and takes action $a_t$ to interact with the environment.
The prompt of this step can also be found in Appendix~\ref{app:selfgoal_prompt}.

\section{Experimental Setup}
\label{sec:Setting}

\subsection{Tasks and Environments}

\begin{wraptable}{r}{0.5\textwidth}
\vspace{-30pt}
\small 
  \centering
    \caption{The categorization of studied tasks.}
    \vspace{0.1cm}
    \begin{tabular}{lccc}
    \toprule
    \textbf{Task} & \textbf{Rounds} & \textbf{Task Type}\\
    \midrule
    Public Goods Game & Single & Competitive\\
    Guess 2/3 of the Average & Single & Cooperative\\
    First-price Auction & Multiple & Competitive\\
    Bargaining & Multiple & Cooperative \\
    \bottomrule
    \end{tabular}
  \label{tab:environments}
\end{wraptable}

We evaluate \method across 4 dynamic tasks with high-level goals, including \textbf{Public Goods Game}, \textbf{Guess 2/3 of the Average}, \textbf{First-price Auction}, and \textbf{Bargaining}, which are implemented by existing works~\citep{huang2024far,chen2023money,lewis-etal-2017-deal}.
As seen in Table~\ref{tab:environments}, they are either single-round or multi-round games, requiring the collaboration or competition of multiple agents.
Note that agents in multi-round games will only receive delayed rewards at the end of the game.
In our experiments, we repeat single-round games for $T=20$ times and multi-round games for $T=10$ times for stable results.

\paragraph{Public Goods Game: GAMA-Bench}
We use \textbf{GAMA-Bench}~\citep{huang2024far} as the implemented environment for this game. 
Specifically, each of $N=5$ players privately decides the number of tokens contributed to a public pot. 
The tokens in the pot are multiplied by a factor $R$ $(1 \leq R \leq N)$, and the created ``public good'' is distributed evenly among all players. 
Players keep any tokens they do not contribute. 
A simple calculation reveals that for each token a player contributes, their net 
gain is $\frac{R}{N}-1$ (i.e., income-contribution).
Since this value is negative, it suggests that the most rational strategy for each player is to contribute no tokens. 
This strategy results in a Nash equilibrium~\citep{daskalakis2009complexity} in the game. 
$N$ agents using the same backbone model and equipped with the same method (\eg, CLIN or \method ) play games with each other to observe group behavior.
Following \cite{huang2024far}, we set $R=2$. 

\paragraph{Guess 2/3 of the Average: GAMA-Bench} 
Using the implementation of \textbf{GAMA-Bench}~\citep{huang2024far}, $N$ players independently choose a number between 0 and 100~\citep{Ledoux1981},
and whoever has the number closest to two-thirds of the group's average wins the game. 
This setup effectively tests players' theory-of-mind (ToM) abilities~\citep{kosinski2023theory,mao2023review}.
In behavioral economics, the Cognitive Hierarchy Model \citep{CFC2004} categorizes players as follows: Level-0 players choose numbers randomly. Level-1 players assume others are Level-0 and pick two-thirds of an expected mean of 50. 
Level-$k$ players believe that the participants include levels $0$ to $k-1$, and therefore choose $(2/3)^k \times 50$. The optimal outcome is to choose 0 for all players, achieving a Nash equilibrium.
In this game, $N=5$ agents using same backbone model with the same prompting method (\eg, \method) play games with each other to observe group behavior.

\paragraph{First-price Auction: AucArena} 
We use \textbf{AucArena}~\citep{chen2023money} as the implementation of first-price auctions.
An auctioneer collects and announces the bids of all participants, revealing the current highest bid. 
Participants must publicly make their decisions after privately considering their bids.
The auction comprises if $K=15$ items with values ranging from \$2,000 to \$10,000, with an increment of \$2,000 between each item.
These items are presented in a randomized sequence, making the auction last for $K=15$ rounds. 
$N=4$ agents participate in the auction as bidders.
Each agent aims to secure the highest profit by the end of the auction and thereby outperform all competitors.
In our experiment, we set the budget for each bidder at \$20,000. We have an agent, enhanced by various methods (\eg, \method), using different backbone models to compete against three identical opponents powered by the same model (GPT-3.5~\citep{openai2022chatgpt}). 

\paragraph{Bargaining: DealOrNotDeal} 
We use \textbf{DealOrNotDeal}~\citep{lewis-etal-2017-deal} to implement the bargaining over multiple issues.
$N=2$ agents, namely Alice and Bob, are presented with sets of items (e.g., books, hats, balls) and must negotiate their distribution.
Each agent is randomly assigned an integer value between 0 and 10 for each item, ensuring that the total value of all items for any agent does not exceed 10. 
The bargaining goes on for $K=10$ rounds, and if the agents fail to agree on the distribution of items within 10 rounds, neither party profits.
The goal is to minimize profit discrepancies between the two agents. 
We randomly select $M=50$ items for Alice and Bob to negotiate over. 
The final profits at the end of the negotiation for Alice and Bob are defined as $P_{Alice}$ and $P_{Bob}$, respectively.
Note that, we alter the prompting methods of the agent behind Alice, and keep Bob fixed (GPT-3.5).

\subsection{Agent Framework Baselines and Backbone LLMs}
We adopt two types of agent frameworks providing guidance for achieving high-level goals in the above tasks.\footnote{Implementation details are in Appendix~\ref{app:implementation}.}
One is \textbf{task decomposition} framework, including ReAct~\citep{yao2023react} and ADAPT~\citep{prasad2024adapt}.
ReAct enables agents to reason before acting, while ADAPT recursively plans and decomposes complex sub-tasks when the LLM cannot execute them.
Another is \textbf{experience summarization} framework, including Reflexion~\citep{shinn2023reflexion} and CLIN~\citep{majumder2023clin}. 
Reflexion prompts agents to reflect on failed task attempts and retry. 
CLIN creates a memory of causal abstractions to assist trials in future by reflecting on past experiences, expressed as \goal{A [may/should] be necessary for B.}.

To drive these language agent frameworks, we use the following LLMs: \textbf{GPT-3.5-Turbo} (gpt-3.5-turbo-1106)~\citep{openai2024gpt4} and \textbf{GPT-4-Turbo} (gpt-4-1106-preview)~\citep{openai2024gpt4};
\textbf{Gemini 1.0 Pro}~\citep{team2023gemini}; \textbf{Mistral-7B-Instruct-v0.2}~\citep{jiang2023mistral} and a Mixture of Experts (MoE) model \textbf{Mixtral-8x7B-Instruct-v0.1}~\citep{jiang2024mixtral};
\textbf{Qwen 1.5} (7B and 72B variants)~\citep{bai2023qwen}.
The temperature is set to 0 to minimize randomness.

\subsection{Metrics for Tasks}
In GAMA-Bench's Public Goods Game~\citep{huang2024far}, where $N$ players participating in repeated $T$ times, 
the score $S_1$ for this game is then given by:
$S_1=\frac{1}{NT} \sum_{i j} C_{i,j}$,
where $C_{i,j} \in [0, 1]$ is the proposed contribution of player $i$ in round $j$.

In GAMA-Bench's Guess 2/3 of the Average Game~\citep{huang2024far}, the score $S_2$ is calculated by 
$S_{2}=100-\frac{1}{NT} \sum_{ij} C_{i,j}$,
where $C_{i,j}$ is the number chosen by player $i$ in round $j$.

In AucArena's First-price Auction~\citep{chen2023money}, we use the TrueSkill Score~\citep{herbrich2006trueskill, minka2018trueskill} (Appendix~\ref{app:trueskill}) to rank the profits of agents. 
TrueSkill Score estimates dynamic skill levels $(\mu)$ through Bayesian statistics while considering the uncertainty $(\sigma)$ in their true skills.
Thus the performance score of an agent is defined as $S_{3} = \text{TrueSkill Score}$. 
This method is commonly used in competitions such as online games or tournaments. 

In DealOrNotDeal's Bargaining Game~\citep{lewis-etal-2017-deal}, 
we calculate the absolute difference in their profits:
$S_{4}=\frac{|P_{Alice} - P_{Bob}|}{M}$,
where $P_{Alice}, P_{Bob}$ represents the profits at the end of the negotiation, and $M$ is the number of items to negotiate on. ($S_{4}$ can also be represented by TrueSkill Score for convenience.)

\section{Results and Analysis}
\label{sec:Results}
\subsection{Main Results}
\label{sec:mainresults}

\newcolumntype{B}{>{\columncolor{blue!4}}c}
\newcolumntype{d}{>{\columncolor{brown!4}}c}
\newcolumntype{q}{>{\columncolor{green!4}}c}

\setlength\tabcolsep{1.5pt}
\begin{table}[t]
\small
  \caption{Comparison of the \method powered by different models with alternative methods across four scenarios. The best results are \textbf{bolded}, and the second best ones are \uline{underlined}. }
  \centering
    \begin{tabular}{lBBBBBqqqqq}
    \toprule
        \multirow{2}{*}{\textbf{Methods}} & \multicolumn{1}{c}{\textbf{ReAct}} & \multicolumn{1}{c}{\textbf{ADAPT}} & \multicolumn{1}{c}{\textbf{Reflexion}} & \multicolumn{1}{c}{\textbf{CLIN}} & \multicolumn{1}{c}{\textbf{\method}} & \multicolumn{1}{c}{\textbf{ReAct}} & \multicolumn{1}{c}{\textbf{ADAPT}} & \multicolumn{1}{c}{\textbf{Reflexion}} & \multicolumn{1}{c}{\textbf{CLIN}} & \multicolumn{1}{c}{\textbf{\method}} \\
        \cmidrule(lr{0.75em}){2-6} \cmidrule(lr{0.75em}){7-11}
        & \multicolumn{5}{c}{\textbf{Public Goods Game: GAMA~\cite{huang2024far} ($S_{1} \downarrow$)}} & \multicolumn{5}{c}{\textbf{Guess 2/3 of the Average: GAMA~\cite{huang2024far} ($S_{2} \uparrow$)}} \\
    \midrule
    Mistral-7B  & 55.70  & 46.00 & 51.28  & \uline{41.00}   & \textbf{28.45} & 89.43  & 84.91 & \uline{92.65}  & 91.95 &\textbf{93.64} \\
    Mixtral-8x7B & 46.05  & 55.80 & \uline{34.65}  & 52.69  & \textbf{32.00} & 82.16  & 79.46 & \textbf{89.73}  & 74.33   & \uline{89.50} \\
    Qwen-7B   & 66.55  & 56.44 & 60.15  & \uline{55.59} & \textbf{54.93} & 65.11 & 55.95 & \uline{69.99} & 64.22  & \textbf{72.99}  \\
    Qwen-72B   & \uline{20.75} & 22.95 & 21.57  & 24.60  & \textbf{8.45} & 78.87  & 88.77 & \uline{91.47}  & 83.65  & \textbf{94.51}  \\
    \midrule
    Gemini Pro & 37.55  & \uline{25.78} & 34.00  & 39.20  & \textbf{19.20} & \textbf{77.90}  & 73.45 & 71.82  & 76.58  & \uline{77.33} \\
    GPT-3.5   & 61.20  & \uline{42.25} & 46.95  & 47.15  & \textbf{42.19}  & 73.44   & 64.14 & \uline{78.75}  & 63.25  & \textbf{83.28} \\
    GPT-4   & 19.55  & \uline{16.70} & 22.90  & 31.35  & \textbf{11.95} & 92.57 & 91.31 & \uline{94.41}  & 90.88  &\textbf{94.54}  \\
    \midrule
    \multirow{2}{*}{\textbf{Methods}} & \multicolumn{1}{c}{\textbf{ReAct}} & \multicolumn{1}{c}{\textbf{ADAPT}} & \multicolumn{1}{c}{\textbf{Reflexion}} & \multicolumn{1}{c}{\textbf{CLIN}} & \multicolumn{1}{c}{\textbf{\method}} & \multicolumn{1}{c}{\textbf{ReAct}} & \multicolumn{1}{c}{\textbf{ADAPT}} & \multicolumn{1}{c}{\textbf{Reflexion}} & \multicolumn{1}{c}{\textbf{CLIN}} & \multicolumn{1}{c}{\textbf{\method}} \\
        \cmidrule(lr{0.75em}){2-6} \cmidrule(lr{0.75em}){7-11}
    & \multicolumn{5}{c}{\textbf{First-price Auction: AucArena~\cite{chen2023money} ($S_{3} \uparrow$)}} & \multicolumn{5}{c}{\textbf{Bargaining: DealOrNotDeal~\citep{lewis-etal-2017-deal}($S_{4} \downarrow$)}} \\
    \midrule
    Mistral-7B & 23.91 & 23.03 & \uline{26.24}  & 24.27  & \textbf{28.21} & 2.57 & 2.38 & \uline{1.97}  & 2.32  & \textbf{1.88} \\
    Mixtral-8x7B & 35.85  & 32.35 & 33.18  & \uline{36.37}  & \textbf{39.23} & 2.38  & 2.66 & 2.46  & \uline{2.34}  & \textbf{1.97} \\
    Qwen-7B  & 29.88  & 30.15 & 32.97  & \uline{33.44} & \textbf{33.50} & 2.83 & 2.88 & 3.15  & \uline{2.73}  & \textbf{2.05} \\
    Qwen-72B  & 34.77 & 34.25 & \uline{35.92}   & 34.24    & \textbf{36.48} & 2.59  &2.10 & \uline{2.06}  & 2.26   & \textbf{2.00} \\
    \midrule
    Gemini Pro & 36.12 & 36.47 & \uline{38.82}  & 36.79   & \textbf{39.28} & \uline{2.10}  & 2.33 &  2.28 & 2.36  & \textbf{1.95} \\
    GPT-3.5  & \uline{22.85}  & 22.10 & 22.00  & 21.21  & \textbf{27.40} & \uline{2.31}  & 2.95 & 2.44  & 2.87  & \textbf{2.20} \\
    GPT-4  & 36.46  & 35.40 & 34.41  & \uline{38.98}  & \textbf{39.02} & 1.94  & \uline{1.80} & 1.92  & 1.83  & \textbf{1.71}  \\
    \bottomrule
    \end{tabular}
  \label{tab:main}
\end{table}

The main results across 4 scenarios are presented in Table~\ref{tab:main}. 
Overall, our \method significantly outperforms all baseline frameworks in various environments containing high-level goals, where larger LLMs produce higher gains. 
When diving into the generated guidelines and corresponding agents' behaviors, we find that some of those subgoals given by task decomposition methods like ReAct and ADAPT are no longer suited for the current situation.
For example, \goal{bid on the most expensive item} is not useful when the budget is tight. 
Moreover, task decomposition before interacting with the environment does not consider the practical experience, leading to broad and meaningless guidance.
For example, in Public Goods Game, ADAPT provides broad subgoals like \goal{It's important to strike a balance between contributing enough tokens to the public pot to earn a significant payoff while retaining enough tokens in my private collection for future rounds}.
In contrast, post-hoc experience summarization methods, i.e., Reflexion and CLIN, tend to induce too detailed guidelines, lacking a correlation with the main goal and might deviating agents from their paths. 
For example, CLIN produces subgoals focusing on minutiae, such as \goal{Considering the distribution of numbers chosen by opponents may be necessary to make an informed decision on your own selection.}

In comparison, \method overcomes both of the shortcomings. 
At each round, \method decomposes new nodes referring to existing guidance, aligning with the main goal as the game progresses. 
For example, in Public Good Game, the initial subgoal is \goal{The player aims to contribute strategically based on their assessment of other players' behaviors and the overall distribution of tokens in the public pot.}
If all players contribute less to the public pot during the game, \method absorbs the observation and refines existing nodes to \goal{If the player notices that the average contribution of the group has been increasing in recent rounds, they might choose to contribute fewer tokens in the current round to avoid over-contributing and potentially losing out on their own gain.}
According to the new subgoal as a practical guideline, agents can dynamically adjust their contributions.\footnote{More details of \tree are in Appendix~\ref{app:goaltree_example}.}

Interestingly, \method shows superior performance in smaller LLMs as well, while others can not due to the deficiency of induction and summarization capability of these models.
For example, CLIN is 0.7 inferior to Reflexion for Mistral-7B and 5.77 for Qwen-7B in Guess 2/3 of the Average, but \method brings improvements consistently.
This can be attributed to the logical, structural architecture of \tree in \method.
At each time for decomposition, the model receives existing subgoals on the last layer of \tree as clear references, making it easy for decomposition.

\setlength{\intextsep}{2pt} 
\begin{wraptable}{r}{0.62\textwidth}
\small 
  \centering
    \caption{Result of auction competitions between the reported five agents with baseline frameworks and our \method.}
    \vspace{0.1cm}
    \begin{tabular}{@{}l@{\hspace{3pt}}c@{\hspace{3pt}}c@{\hspace{3pt}}c@{\hspace{3pt}}c@{\hspace{3pt}}c@{}} 
    \toprule
        \textbf{Methods} & \textbf{ReAct} & \textbf{ADAPT} & \textbf{Reflexion} & \textbf{CLIN} & \textbf{\method} \\
    \midrule
    GPT-3.5  & $23.96_{\pm 1.72}$ & $20.46_{\pm 1.79}$  & \uline{$25.72_{\pm 1.71}$}  & $22.95_{\pm 1.73}$  & $\mathbf{29.59_{\pm 1.99}}$  \\
    GPT-4  & $22.62_{\pm 1.80}$  &  $24.85_{\pm 1.78}$ & $21.79_{\pm 1.79}$   & \uline{$27.16_{\pm 1.74}$}   & $\mathbf{28.98_{\pm 1.88}}$ \\
    \bottomrule
    \end{tabular}
  \label{tab:auction}
\end{wraptable}

\paragraph{Competition between Different Agent Framework}
Previous results are mostly evaluated against a fixed baseline (GPT-3.5).
To understand how these agent frameworks behave when competing with each other, we set an AucArena for a multi-agent comparison.
As shown in Table~\ref{tab:auction}, \method has a clear advantage over baselines.
When looking closer at the bidding behaviors, we find that other methods tend to be overly cautious. 
They often stop bidding or avoid participating once bidding starts, resulting in zero profits. 
However, \method takes a different approach by bidding frequently in the early stages of the bidding war when competition is less intense. 
This allows for purchasing high-priority items early on, avoiding fierce competition in the later stages.

\subsection{Analysis of \method}
\label{sec:analysis}

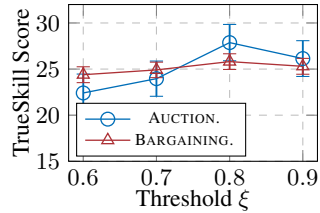
\begin{wrapfigure}{r}{0.35\linewidth}
    \centering
    \small
    \pgfplotsset{width=1\linewidth,height=0.75\linewidth,compat=1.8}
\begin{tikzpicture}
\begin{axis}[
    xmin=0.58, xmax=0.92,
    ymin=15, ymax=32,
    xtick={0.6, 0.7, 0.8, 0.9},
    ytick={10, 15, 20, 25, 30},
    legend pos=south west,
    ymajorgrids=true,
    xmajorgrids=true,
    grid style=dashed,
    xlabel={Threshold $\xi$},
    ylabel={TrueSkill Score},
    x label style={at={(axis description cs:0.5,-0.125)},anchor=north},
    y label style={at={(axis description cs:-0.125,0.5)},anchor=south},
    legend style={nodes={scale=0.7, transform shape}}
]
\addplot[
    color=NavyBlue,
    mark=o,
    line width=0.5pt,
    mark size=2.6pt,
    error bars/.cd,
    y dir=both, y explicit,
    error bar style={line width=0.7pt, color=NavyBlue},
    error mark options={rotate=90, NavyBlue, mark size=2.5pt}
    ]
    coordinates {
    (0.6, 22.41) +- (0.6, 2.05)
    (0.7, 23.96) +- (0.7, 1.91)
    (0.8, 27.87) +- (0.8, 1.97)
    (0.9, 26.14) +- (0.9, 1.94)
    };
    \addlegendentry{\textsc{Auction.}}

\addplot[
    color=Maroon,
    mark=triangle,
    line width=0.5pt,
    mark size=2.6pt,
    error bars/.cd,
    y dir=both, y explicit,
    error bar style={line width=0.7pt, color=Maroon},
    error mark options={rotate=90, Maroon, mark size=2.5pt}
    ]
    coordinates {
    (0.6, 24.39) +- (0.6, 0.86)
    (0.7, 24.91) +- (0.7, 0.84)
    (0.8, 25.81) +- (0.8, 0.85)
    (0.9, 25.30) +- (0.9, 0.86)
    };
    \addlegendentry{\textsc{Bargaining.}}
\end{axis}
\end{tikzpicture}
    \vspace{-0.2cm}
    \caption{Granularity control of the threshold $\xi$ in \method's stopping mechanism.}
    \label{fig:stopping}
\end{wrapfigure}

\paragraph{How does the granularity of guidelines in \tree affect task solving?}

As discussed in $\mathsection$\ref{sec:mainresults}, \method adjusts to the dynamic environment by setting different depths, where subgoal nodes of deeper layers provide more detailed instructions. 
Here, we explore how such granularity affects the performance of \method.
We use Auction and Bargaining environments as testbeds, and modify the level of subgoals by setting the threshold $\xi$ in the stopping mechanism as 0.6, 0.7, 0.8, and 0.9.
According to Figure~\ref{fig:stopping}, the agent's performance initially improves with increasing depth but eventually diminishes. 
A shallow tree ($\xi=0.6$) lacks guidance details, thus leading to the poorest performance. 
Yet, the deepest tree ($\xi=0.9$) does not show superior performance, probably because repetitive guidance interferes with model selection of useful guidance.
Redundant nodes increase the candidate set, making it difficult for the search module to select all the valuable nodes.
In fact, the search module always focuses on multiple nodes representing the same meaning, resulting in the loss of other helpful nodes.
This experiment confirms that more detailed instructions help language agents achieve high-level goals, but only with a balanced, adaptive depth of the guidance tree to mitigate the drawbacks of overly detailed guidance.

\begin{wrapfigure}{r}{0.35\linewidth}
    \centering
    \small
    \pgfplotsset{width=1\linewidth,height=0.85\linewidth,compat=1.8}
\footnotesize

\begin{tikzpicture}
    \begin{axis}[
        ybar,
        bar width=8pt,
        xtick distance=1,
        symbolic x coords={\pac, \nac},
        xticklabels={a,Auction,Bargaining},
        ylabel=TrueSkill Score,
        ymin=0, ymax=32,
        enlarge x limits=0.45,
        scaled ticks=false,
        legend pos=south east,
        xtick style={
            /pgfplots/major tick length=0pt,
        },
        legend style={nodes={scale=0.8, transform shape},font=\footnotesize}
    ]
        \addplot+ [
            color=Maroon!40,
            draw=Maroon,
            error bars/.cd,
            y dir=both,
            y explicit,
            error mark options={
              pink,
              mark size=0.2pt,
              line width=4pt
            }
        ] coordinates {
            (\pac,23.93) +- (2.23,2.23)
            (\nac,25.06) +- (1.01,1.01)
        };

        \addplot+ [
            color=BlueGreen!30,
            draw=BlueGreen,
            error bars/.cd,
            y dir=both,
            y explicit, 
            error mark options={
              BlueGreen,
              mark size=0.2pt,
              line width=4pt
            }
        ] coordinates {
            (\pac,24.24) +- (2.32,2.32)
            (\nac,23.85) +- (0.98,0.98)
        };

        \addplot+ [
            color=NavyBlue!30,
            draw=NavyBlue,
            error bars/.cd,
            y dir=both,
            y explicit, 
            error mark options={
              Blue,
              mark size=0.2pt,
              line width=4pt
            }
        ] coordinates {
            (\pac,27.68) +- (2.25,2.25)
            (\nac,26.45) +- (1.02,1.02)
        };
        
        \legend{
            \text{Random Selection},
            \text{Embedding Similarity},
            \text{LLM-based Search.},
        }
    \end{axis}
\end{tikzpicture}
    \vspace{-0.2cm}
    \caption{Ablation study of different search modules.}
    \label{fig:searcher}
\end{wrapfigure}
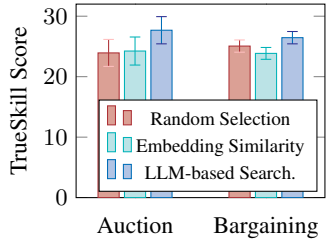

\paragraph{Ablation Study of Search Module}

\textit{Can the Search Module in \method succeed in finding useful subgoal nodes?}
We employ two methods as baselines to replace the original LLM-based search module, which is instantiated with GPT-3.5.
One baseline is \textit{random selection}, where we randomly choose an node from the set of subgoal nodes. 
The other is the selection based on \textit{embedding similarity}, which selects the subgoals most similar to the current situation based on cosine similarity. 
On multi-round games as Auction and Bargaining, we keep the Trueskill Score for evaluating the rankings of these methods.
As shown in Figure~\ref{fig:searcher}, the LLM search module gains a better score in both games. 
Besides, similarity-based method performs worse than random selection in Bargaining, which could be the reason that the guidance is usually short, making it hard to capture semantic embeddings between subgoals and situations.
This experiment demonstrates the rationality of the LLM-based search module in \method's design.

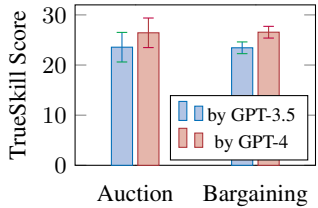
\begin{wrapfigure}{r}{0.38\linewidth}
    \centering
    \small
    \pgfplotsset{width=0.9\linewidth,height=0.7\linewidth,compat=1.5}

\begin{tikzpicture}
    \begin{axis}[
        ybar,
        bar width=8pt,
        xtick distance=1,
        symbolic x coords={\pac, \nac},
        xticklabels={a,Auction,Bargaining},
        ylabel=TrueSkill Score,
        ymin=0, ymax=32,
        enlarge x limits=0.5,
        scaled ticks=false,
        legend pos=south east,
        xtick style={
            /pgfplots/major tick length=0pt,
        },
        legend style={nodes={scale=0.8, transform shape},font=\footnotesize}
    ]
        \addplot+ [
            color=NavyBlue!30,
            draw=NavyBlue,
            error bars/.cd,
            y dir=both,
            y explicit,
            error mark options={
              Green,
              mark size=0.2pt,
              line width=4pt
            }
        ] coordinates {
            (\pac,23.56) +- (2.95,2.95)
            (\nac,23.44) +- (1.17,1.17)
        };

        \addplot+ [
            color=Maroon!30,
            draw=Maroon,
            error bars/.cd,
            y dir=both,
            y explicit, 
            error mark options={
              purple,
              mark size=0.2pt,
              line width=4pt
            }
        ] coordinates {
            (\pac,26.43) +- (2.95,2.95)
            (\nac,26.55) +- (1.17,1.17)
        };

        \legend{
            by GPT-3.5,
            by GPT-4,
        }
    \end{axis}
\end{tikzpicture}
    \vspace{-0.2cm}
    \caption{
    Ablation study of the model that generates \tree, either by a stronger (GPT-4) or weaker (GPT-3.5) model. The rest of the agent framework is driven by GPT-3.5.}
    \label{fig:decomposer}
\end{wrapfigure}

\paragraph{How does the quality of \tree affect goal achievement?}
To explore the influence of \tree on \method, we conduct an experiment in Auction and Bargaining Games by replacing the model that constructs \tree with GPT-4 or GPT-3.5 for comparison, while keeping the model that utilizes the tree fixed as GPT-3.5. 
Results in Figure~\ref{fig:decomposer} illustrate that higher-quality \tree (from GPT-4) significantly boosts the performance of \method, with gains of +2.87 in Auction and +3.10 in Bargaining compared to one using GPT-3.5. 
This improvement comes from more abundant and higher-quality guidance, generated by a strong model equipped with better understanding and summarizing capabilities. (We also conduct an ablation study about the influence of pruning on \tree in Appendix~\ref{app:pruning}.)

\begin{figure*}[t]
    \centering
    \includegraphics[width=0.8\linewidth]{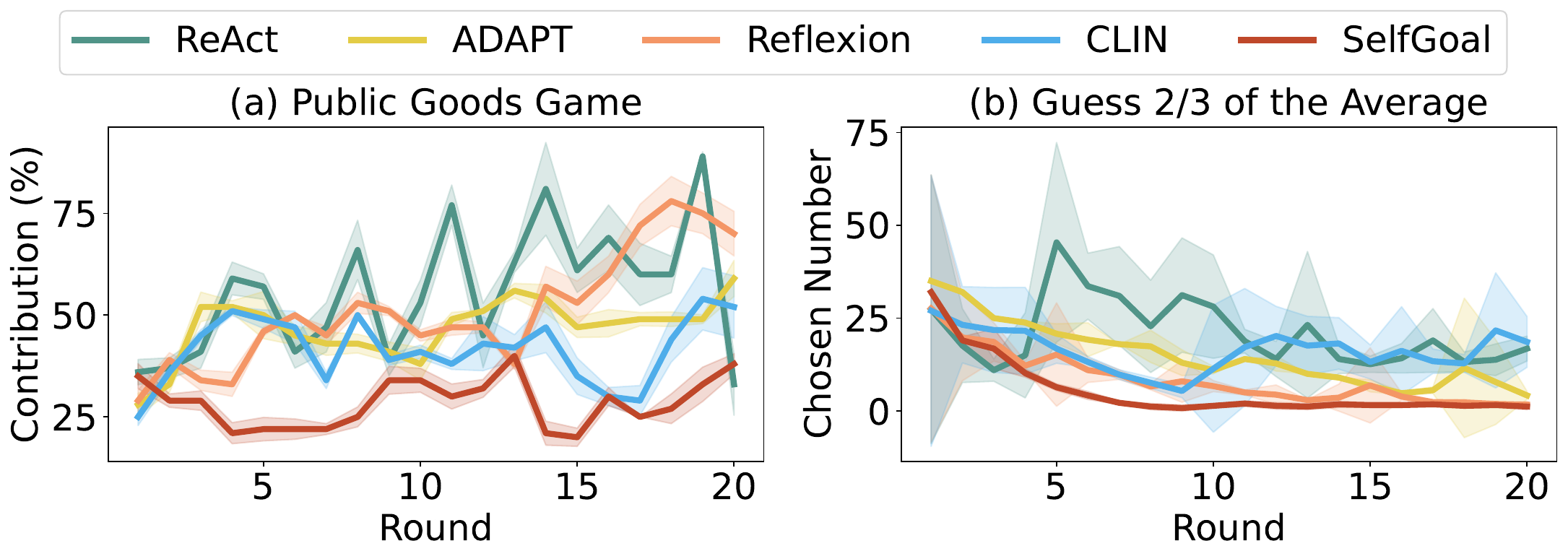}
    \vspace{-0.2cm}
    \caption{Patterns of model behavior in repeated games. (a): Fluctuations in contributions within the Public Goods game. The agent equipped with \method displays more rational behavior (\ie, achieving a Nash equilibrium) by consistently contributing fewer tokens than other methods.
    (b): Adjustments in number predictions within the Guessing Game. Our \method shows enhanced ToM abilities by converging to a guess of zero more quickly in each round.
    }
    \label{fig:behavior}
    \vspace{-0.2cm}
\end{figure*}

\paragraph{Can \method improve the rationality in agents' behaviors?}

Aside from the final performance gain, we are also interested in whether each agent behavior at every turn benefits from \method.
Therefore, we use two games from GAMA-Bench to examine the impact of \method on model behavior, where behavioral changes are easier to evaluate.
Here, we use LLMs with great improvement from \method, i.e., Mistral-7B for Public Goods Game and Qwen-72B for Guessing 2/3 Average Number Game.
We record patterns in the model's number predictions and token contributions by visualizing data from 20 repeated experiments. 
Note that \tree is updated across these 20 rounds of games.
With \method, agents in the Public Goods scenario consistently act more rationally compared to those using alternative methods, as illustrated in Figure~\ref{fig:behavior}(a). 
For the Guessing Game, enhanced models showed smoother, more steadily declining curves, indicating quicker convergence to the Nash equilibrium, as depicted in Figure~\ref{fig:behavior}(b). 

\setlength\tabcolsep{3pt}
\begin{figure*}[t]
    \centering
    \includegraphics[width=0.95\linewidth]{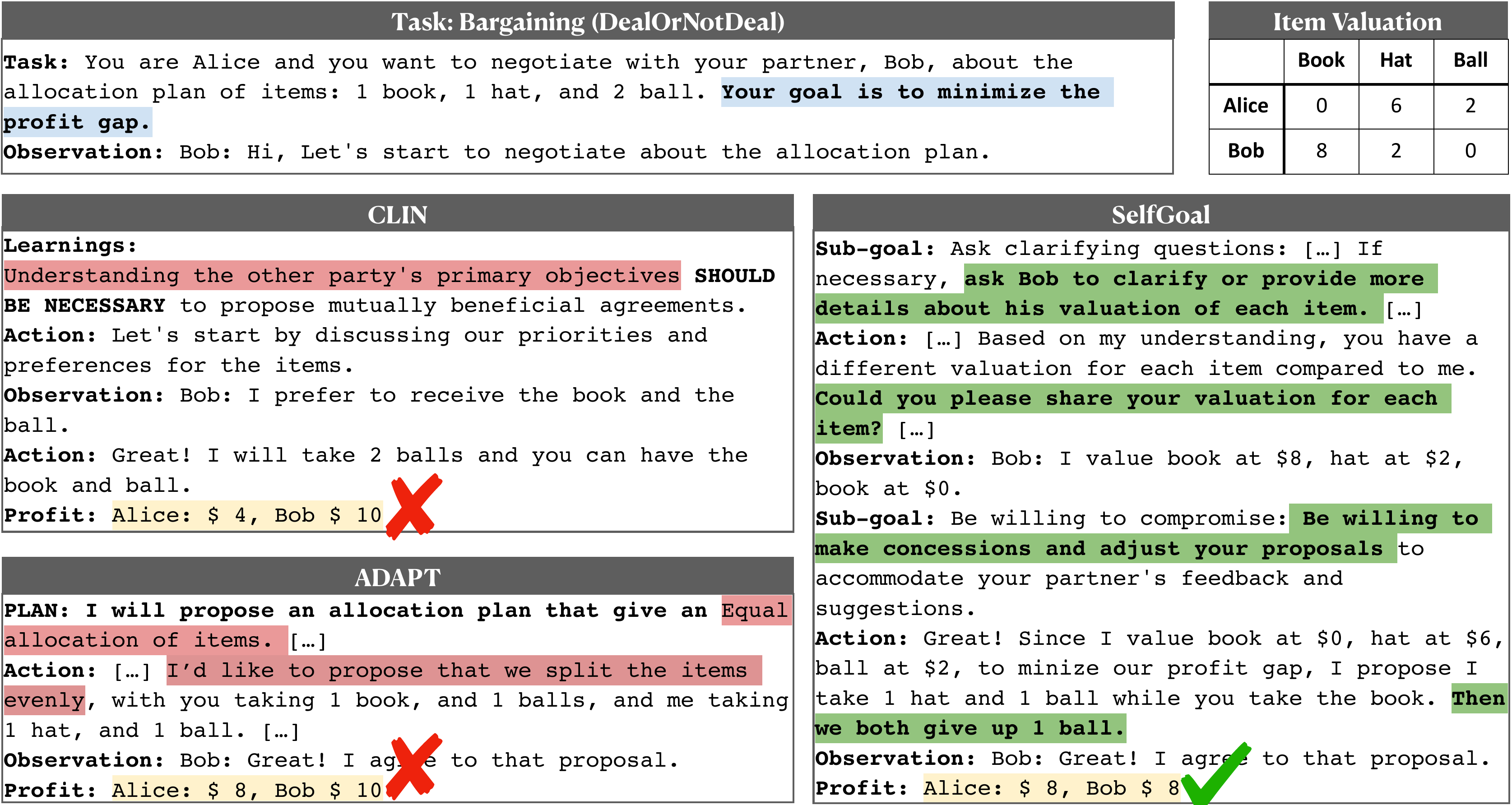}
    \caption{In the Bargaining task, Mistral-7B with CLIN or ADAPT gives guidance that is either too broad or too detailed resulting in large profit discrepency, whereas \method is successful.
    }
    \label{fig:case_study}
\end{figure*}

\subsection{Case Study}
To illustrate how agents from different frameworks reason and plan in a dynamic environment, we conduct a case study using Mistral-7B, a small LLM, as the backbone in a bargaining game (Figure~\ref{fig:case_study}). 
We find that \method's emphasis on granularity control offers clear advantages. 
\method provides agents with actionable guidance such as \goal{ask clarifying questions}, prompting agents to pay early attention to their opponent's psychological assessment and different valuations of items. 
After acquiring a partner's valuation, \method then gives guidance such as \goal{make concessions}, leading the agent to propose a plan that gives up a particular item in exchange for minimizing the profit difference.

In contrast, CLIN advises agents to \goal{consider the preference of the partner}, which leads agents to focus on the opponent's preferences but may result in plans that sacrifice their own interests to improve the other party's income. 
ADAPT, which decomposes tasks beforehand, provides very broad advice such as \goal{equal allocation}. 
This generic advice aims to minimize the profit gap but may not be suitable for scenarios lacking knowledge of the partner's valuation. 
Consequently, the model proposes allocation plans without first clarifying the partner's valuations, assuming that all participants have the same valuation for each item.

\section{Conclusion}
\label{sec:Conclusion}
In this paper, we introduce \method, an agent framework that enhances the capabilities of LLMs for achieving high-level goals across various dynamic tasks and environments.
We demonstrate that \method significantly improves agent performance by dynamically generating and refining a hierarchical \tree of contextual subgoals based on interactions with the environments. 
Experiments show that this method is effective in both competitive and cooperative scenarios, outperforming baseline approaches. 
Moreover, \tree can be continually updated as agents with \method further engage with the environments, enabling them to navigate complex environments with greater precision and adaptability. 
However, we also notice that although \method is effective for small models, there is still a demand for the understanding and summarizing capability of models, which might prevent \method from achieving its full effectiveness.

\bibliography{tree}
\bibliographystyle{unsrt}

\clearpage
\appendix

\section{\method Details}
\subsection{Does pruning the \tree affect search quality?}
\label{app:pruning}
\setlength{\intextsep}{1pt} 
\begin{wraptable}{r}{0.45\textwidth}
\centering
\caption{Comparison of agents guided by \tree with and without pruning.}
\small
\vspace{0.2cm}
\begin{tabular}{lcc}
\toprule
\multirow{2}{*}{\tree \textbf{\quad}}  & \multicolumn{2}{c}{\textbf{Scenario}}    \\ \cmidrule{2-3} 
& \textbf{Auction} & \textbf{Bargaining} \\ 
\midrule
Pruned                        & $24.74 \pm 3.22$	& $24.90 \pm 1.21$ \\
w/o Pruned        & $\mathbf{25.25 \pm 3.23}$	& $\mathbf{25.09 \pm 1.21}$ \\
\bottomrule
\end{tabular}

\label{tab:prune}
\end{wraptable}

We investigate whether pruning nodes not selected for a long time from the target tree affects the Search Module's decisions. 
Pruning begins after the Decompose Module completes building the tree, and nodes unselected for more than five consecutive rounds will be deleted.
We assess the impact of pruning on GPT-3.5's performance in Auction and Bargaining. 
As shown in Table~\ref{tab:prune}, the TrueSkill Score with and without pruning are similar. 
This suggests that nodes not chosen for extended periods do not compromise the Search Module's decision-making effectiveness. 
This efficiency likely results from our Search Module using prior knowledge from LLM to identify and avoid selecting unnecessary nodes, akin to lazy deletion.
For efficiency, these redundant nodes are also removed every five rounds.

\subsection{Implementation Details}
\label{app:implementation}
We compare our \method with the following methods: ReAct~\citep{yao2023react}, which induces an LLM actor to engage in preliminary reasoning about the task before initiating action, Reflexion~\citep{shinn2023reflexion}, which encourages an LLM actor to re-assess unsuccessful task attempts before attempting the task again, CLIN~\citep{majumder2023clin}, which leverages historical insights to deduce transition strategies, articulated as ``A [may/should] be necessary for A''.
To adapt these methods to our experimental environment, we update the memory of the CLIN/Reflexion approach at each timestep within a single trial, whether it is a bid in the Auction environment, a dialogue round in the Negotiation environment, or a game round in GAMA-Bench. 
Specifically, for Reflexion, the model uses historical steps from the current trial to generate verbal self-reflections. 
These self-reflections are then added to long-term memory, providing valuable feedback for future trials. In the case of CLIN, we use the BASE method due to the absence of a training set in our environment. 
The memory is updated at each step by prompting the model with historical steps from the current trial and all previous memories to generate an updated memory, which includes a new list of semi-structured causal abstractions. 
This updated memory is then incorporated into the historical memories.

\subsection{Details of TrueSkill Score} 
\label{app:trueskill}
In a game with a population of $n$ players $\{1, \ldots, n\}$, consider a match where $k$ teams compete. The team assignments are specified by $k$ non-overlapping subsets $A_j \subset\{1, \ldots, n\}$ of the player population, with $A_i \cap A_j=\emptyset$ for $i \neq j$. The outcome $\mathbf{r}:=\left(r_1, \ldots, r_k\right) \in\{1, \ldots, k\}$ is defined by a rank $r_j$ for each team $j$, with $r=1$ indicating the winner and draws possible when $r_i=r_j$. Ranks are based on the game's scoring rules.

The probability $P(\mathbf{r} \mid \mathbf{s}, A)$ of the game outcome $\mathbf{r}$ is modeled given the skills $\mathbf{s}$ of the participating players and the team assignments $A:=\left\{A_1, \ldots, A_k\right\}$. From Bayes' rule, we get the posterior distribution
$$
p(\mathbf{s} \mid \mathbf{r}, A)=\frac{P(\mathbf{r} \mid \mathbf{s}, A) p(\mathbf{s})}{P(\mathbf{r} \mid A)} .
$$
We assume a factorizing Gaussian prior distribution, $p(\mathbf{s}):=\prod_{i=1}^n {N}\left(s_i ; \mu_i, \sigma_i^2\right)$. Each player $i$ is assumed to exhibit a performance $p_i \sim {N}\left(p_i ; s_i, \beta^2\right)$ in the game, centered around their skill $s_i$ with fixed variance $\beta^2$. 

The performance $t_j$ of team $j$ is modeled as the sum of the performances of its members, $t_j:=\sum_{i \in A_j} p_i$. Teams are reordered in ascending order of rank, $r_{(1)} \leq r_{(2)} \leq \cdots \leq r_{(k)}$. Disregarding draws, the probability of a game outcome $\mathbf{r}$ is modeled as
$$
P\left(\mathbf{r} \mid\left\{t_1, \ldots, t_k\right\}\right)=P\left(t_{r_{(1)}}>t_{r_{(2)}}>\cdots>t_{r_{(k)}}\right)
$$
In other words, the order of performances determines the game outcome. If draws are allowed, the winning outcome $r_{(j)}<r_{(j+1)}$ requires $t_{r_{(j)}}>t_{r_{(j+1)}}+\varepsilon$ and the draw outcome $r_{(j)}=r_{(j+1)}$ requires $\left|t_{r_{(j)}}-t_{r_{(j+1)}}\right| \leq \varepsilon$, where $\varepsilon>0$ is a draw margin calculated from the assumed probability of a draw. ${ }^1$

To report skill estimates after each game, we use an online learning scheme called Gaussian density filtering. The posterior distribution is approximated to be Gaussian and is used as the prior distribution for the next game. If skills are expected to change over time, a Gaussian dynamics factor ${N}\left(s_{i, t+1} ; s_{i, t}, \gamma^2\right)$ can be introduced, leading to an additive variance component of $\gamma^2$ in the subsequent prior.

Consider a game with $k=3$ teams with team assignments $A_1=\{1\}, A_2=\{2,3\}$ and $A_3=\{4\}$. Assume that team 1 wins and teams 2 and 3 draw, i.e., $\mathbf{r}:=(1,2,2)$. 
The function represented by a factor graph in our case, the joint distribution $p(\mathbf{s}, \mathbf{p}, \mathbf{t} \mid \mathbf{r}, A)$, is given by the product of all the potential functions associated with each factor. The structure of the factor graph provides information about the dependencies of the factors involved and serves as the foundation for efficient inference algorithms. 
Referring back to Bayes' rule, the quantities of interest are the posterior distribution $p\left(s_i \mid \mathbf{r}, A\right)$ over skills given game outcome $\mathbf{r}$ and team assignments $A$. The $p\left(s_i \mid \mathbf{r}, A\right)$ are calculated from the joint distribution by integrating out the individual performances $\left\{p_i\right\}$ and the team performances $\left\{t_i\right\}$:
$$
p(\mathbf{s} \mid \mathbf{r}, A)=\int_{-\infty}^{\infty} \cdots \int_{-\infty}^{\infty} p(\mathbf{s}, \mathbf{p}, \mathbf{t} \mid \mathbf{r}, A) d \mathbf{p} d \mathbf{t} .
$$

\subsection{Instruction Prompt Examples}
\label{app:selfgoal_prompt}
The instruction prompts of three modules in \method are presented in Listing~\ref{listing:prompt}.
\lstset{
    backgroundcolor=\color[RGB]{245,245,244},
    breaklines=true,
    breakindent=0pt,
    basicstyle=\ttfamily\small,
    emph={Decomposition, Search, Task, Solving, Instruction},
    emphstyle={\bfseries\color{brown}}
}
\begin{lstlisting}[caption={The instruction prompts in \method.},label=listing:prompt]
Decomposition Instruction:

# Main Goal
Humans exhibit numerous behaviors and sub-goals, which can be traced back to the primary aim of survival. For instance:
1. Food Acquisition: To maintain physical and mental functionality, individuals seek nourishment. They target foods with high energy and nutritional values to augment their health, thus enhancing survival possibilities.
2. Shelter Construction: Safe and secure housing is a fundamental human need. It offers protection from potentially harmful natural elements and potential threats.

Imagine you are an agent in an ascending-bid auction. You will compete against other bidders in a bidding war. The price steadily increases as bidders progressively pull out. Eventually, a single bidder emerges as the winner, securing the item at the final bid.

Taking analogy from human behaviors, if your fundamental objective in this auction is "{goal}", what sub-goals you might have?

------------------------------

# Sub-Goal
For the goal: "{sub_goal}", can you further run some deduction for fine-grained goals or brief guidelines?


Search Instruction:

Here's the current scenario:
{scene}
------------------------------
To better reach your main goal: {objective}, in this context, please do the following:
1.Evaluate how the sub-goals listed below can assist you in reaching your main goal given the present circumstances.
Sub-goals:
{guidance}
2. Select {width} most useful sub-goals that will help you reach your
main goal in the current situation, and note their IDs.
Start by explaining your step-by-step thought process. Then, list the {width} IDs you've chosen, using the format of this example: {{"IDs": [1, 3, 10, 21, 7]}}.

Task Solving Instruction:
Here is the current scenarios:

{scene}

------------------------------
Here are some possible subgoals and guidance derived from your primary objective {main_goal}:

{sub_goals}

In this round, You may target some of these subgoals and detailed guidance to improve your strategy and action, to achieve your primary objective.

\end{lstlisting}
We implemented CLIN and Reflexion methods in our environments as presented in Listing~\ref{listing:baseline}.
\lstset{
    backgroundcolor=\color[RGB]{245,245,244},
    breaklines=true,
    breakindent=0pt,
    basicstyle=\ttfamily\small,
    emph={REFLEXION, CLIN, Instruction},
    emphstyle={\bfseries\color{brown}}
}\begin{lstlisting}[caption={The instructions for Reflexion and CLIN.},label=listing:baseline]
REFLEXION Instruction:

You are an advanced reasoning agent that can improve based on self
refection.
Review and reflect on the historical data provided from a past
auction.
{past_auction_log}
Based on the auction log, in a few sentences, diagnose a possible reason for failure or phrasing discrepancy and devise a new, concise, high level plan that aims to mitigate the same failure. Use complete sentences.

CLIN Instruction:

Review and reflect on the historical data provided from a past
auction.
{past_auction_log}
Here are your past learnings:
{past_learnings}
Based on the auction log, formulate or update your learning points that could be advantageous to your strategies in the future. Your learnings should be strategic, and of universal relevance and practical use for future auctions. Consolidate your learnings into a concise numbered list of sentences.
Each numbered item in the list can ONLY be of the form:
X MAY BE NECCESSARY to Y.
X SHOULD BE NECCESSARY to Y.
X MAY BE CONTRIBUTE to Y.
X DOES NOT CONTRIBUTE to Y.

\end{lstlisting}

\subsection{Examples of GoalTree}
\label{app:goaltree_example}
Here, we provide examples of \tree from four environments in Listing~\ref{listing:tree_examples}, with their main goals as follows:
\begin{itemize}
    \item \textbf{Public Goods}: maximize your total token count by the end of the game;
    \item \textbf{Guess 2/3 of the Average}: choose a number that you believe will be closest to 2/3 of the average of all numbers chosen by players, including your selection;
    \item \textbf{First-price Auction}: secure the highest profit at the end of this auction, compared to all other bidders;
    \item \textbf{Bargaining}: minimize the profit gap between yourself and your partner in this negotiation, regardless of your own profit.
\end{itemize}
\lstset{
    backgroundcolor=\color[RGB]{245,245,244},
    breaklines=true,
    breakindent=0pt,
    basicstyle=\ttfamily\small,
    emphstyle={\bfseries\color{brown}}
}\begin{lstlisting}[caption={Examples of \tree in \method.},label=listing:tree_examples]
Public Goods Game:

root: Maximize your total token count by the end of the game.
root-0: Maximizing Contribution
root-0-0: Assess the Current State
root-0-0-2: Long-term Token Accumulation
root-0-0-2-3: Collaboration and Competition
root-0-0-2-3-0: Observation and Analysis
root-0-0-2-3-0-1: Identify Potential Collaborators
root-0-0-2-3-0-1-1: Observe Consistency
root-0-0-2-3-0-1-1-1: Establish Trustworthy Partnerships
root-0-0-2-3-0-1-1-1-2: Monitor Trustworthiness
root-0-0-2-3-0-1-1-1-2-1: Identify Unreliable Contributors
root-0-0-2-3-0-1-1-1-2-1-0: Track and Analyze Contributions
root-0-0-2-3-0-1-1-1-2-1-0-1: Identify Inconsistent Contributors
root-0-0-2-3-0-1-1-1-2-1-0-1-1: Monitor Reliability
root-0-0-2-3-0-1-1-1-2-1-0-1-2: Consider Communication
root-0-0-2-3-0-1-1-1-2-1-0-1-3: Adjust Your Strategy
root-0-0-2-3-0-1-1-1-2-1-0-1-3-2: Anticipate Player Behavior
root-0-0-2-3-0-1-1-1-2-1-0-1-3-4: Risk Management
root-0-0-2-3-0-1-1-1-2-1-0-1-4: Collaborate with Consistent Contributors
root-0-0-2-3-0-1-1-1-2-1-0-1-4-0: Identify Reliable Contributors
root-0-0-2-3-0-1-1-1-2-1-0-1-4-1: Establish Communication
root-0-0-2-3-0-1-1-1-2-1-0-1-4-1-2: Observe Behavioral Patterns
root-0-0-2-3-0-1-1-1-2-1-0-1-4-1-3: Formulate a Joint Strategy
root-0-0-2-3-0-1-1-1-2-1-0-1-4-1-3-1: Optimal Contribution Levels
root-0-0-2-3-0-1-1-1-2-1-0-1-4-1-3-2: Establish Communication
root-0-0-2-3-0-1-1-1-2-1-0-1-4-1-3-3: Adaptation and Flexibility
root-0-0-2-3-0-1-1-1-2-1-0-1-4-1-3-4: Trust and Collaboration
root-0-0-2-3-0-1-1-1-2-1-0-1-4-3: Monitor Consistency
root-0-0-2-3-0-1-1-1-2-1-0-4: Communication and Collaboration
root-0-0-2-3-0-1-1-1-2-1-0-4-2: Encourage Consistency
root-0-0-2-3-0-1-1-1-2-1-0-4-3: Form Alliances
root-0-0-2-3-0-1-1-1-2-1-0-4-3-1: Establish Communication
root-0-0-2-3-0-1-1-1-2-1-0-4-3-2: Coordinate Contribution Efforts
root-0-0-2-3-0-1-1-1-2-1-0-4-3-3: Build Trust and Reliability
root-0-0-2-3-0-1-1-1-2-1-0-4-4: Monitor and Adapt
root-0-0-2-3-0-1-1-1-2-1-2: Communicate and Negotiate
root-0-0-2-3-0-1-1-1-2-1-2-0: Analyze Contribution Patterns
root-0-0-2-3-0-1-1-1-2-1-2-3: Monitor Trustworthiness
root-0-0-2-3-0-1-1-1-2-1-2-4: Adapt to Changing Dynamics
root-0-0-2-3-0-1-1-1-2-1-2-4-1: Form Alliances
root-0-0-2-3-0-1-1-1-2-1-2-4-4: Long-term Planning
root-0-0-2-3-0-1-1-1-2-1-2-4-4-0: Assess the Current Trend
root-0-0-2-3-0-1-1-1-2-1-2-4-4-4: Flexibility in Strategy
root-0-0-2-3-0-1-1-1-2-1-2-4-4-5: Consistency in Contributions
root-0-0-2-3-0-1-1-1-2-1-4: Build a Reputation
root-0-0-2-3-0-1-1-1-2-1-4-2: Observation and Adaptation
root-0-0-2-3-0-1-1-1-2-1-4-4: Communication and Collaboration
root-0-0-2-3-0-1-1-1-2-2: Establish Collaborative Partnerships
root-0-0-2-3-0-1-1-1-2-2-0: Identify Trustworthy Players
root-0-0-2-3-0-1-1-1-2-2-0-2: Consider Long-Term Behavior
root-0-0-2-3-0-1-1-1-2-2-0-2-1: Identify Trustworthy Players
root-0-0-2-3-0-1-1-1-2-2-0-2-3: Adjust Your Strategy
root-0-0-2-3-0-1-1-1-2-2-0-3: Form Alliances
root-0-0-2-3-0-1-1-1-2-2-0-3-1: Assess Trustworthiness
root-0-0-2-3-0-1-1-1-2-2-0-3-3: Mutual Benefit
root-0-0-2-3-0-1-1-1-2-2-0-3-4: Long-Term Collaboration
root-0-0-2-3-0-1-1-1-2-2-0-4: Monitor Changes
root-0-0-2-3-0-1-1-1-2-2-1: Initiate Communication
root-0-0-2-3-0-1-1-1-2-2-2: Reciprocate Trust
root-0-0-2-3-0-1-1-1-2-2-4: Adaptability
root-0-0-2-3-0-1-1-1-2-2-4-0: Assess Other Players' Contributions
root-0-0-2-3-0-1-1-1-2-2-4-2: Identify Potential Alliances
root-0-0-2-3-0-1-1-1-4: Long-term Planning
root-0-0-2-3-0-1-1-1-4-2: Encourage Cooperative Behavior
root-0-0-2-3-0-1-1-1-4-2-0: Establish Trust
root-0-0-2-3-0-1-1-1-4-2-1: Strategic Communication
root-0-0-2-3-0-1-1-1-4-2-1-2: Highlight Long-Term Benefits
root-0-0-2-3-0-1-1-1-4-2-1-3: Negotiate Contribution Strategies
root-0-0-2-3-0-1-1-1-4-2-1-4: Foster Trust and Collaboration
root-0-0-2-3-0-1-1-1-4-2-2: Highlight Mutual Gains
root-0-0-2-3-0-1-1-1-4-2-3: Foster Collaboration
root-0-0-2-3-0-1-1-1-4-2-4: Long-Term Perspective
root-0-0-2-3-0-1-1-1-4-3: Monitor and Adapt
root-0-0-2-3-0-1-1-1-4-3-1: Build Sustainable Partnerships
root-0-0-2-3-0-1-1-1-4-3-3: Strategic Observation
root-0-0-2-3-0-1-1-1-4-3-4: Long-term Adaptation
root-0-0-2-3-0-1-1-1-4-4: Evaluate Long-Term Gains
root-0-0-2-3-0-1-1-1-4-4-2: Monitor Contribution Trends
root-0-0-2-3-0-1-1-2: Monitor Changes in Contributions
root-0-0-2-3-0-1-1-2-2: Form Partnerships
root-0-0-2-3-0-1-1-2-2-1: Establish Communication
root-0-0-2-3-0-1-1-2-2-2: Form Strategic Alliances
root-0-0-2-3-0-1-1-2-2-4: Maximize Collective Gain
root-0-0-2-3-0-1-1-2-3: Anticipate Changes
root-0-0-2-3-0-1-1-2-4: Evaluate Risk-Reward Ratio
root-0-0-2-3-0-1-3: Build Trust and Cooperation
root-0-0-2-3-0-1-4: Monitor Results
root-0-0-2-3-0-1-4-1: Assess Impact on Public Good Payoff
root-0-0-2-3-0-1-4-1-1: Evaluate Public Pot Growth
root-0-0-2-3-0-1-4-1-3: Identify Collaborative Strategies
root-0-0-2-3-0-1-4-1-4: Predict Future Payoff Trends
root-0-0-2-3-0-1-4-2: Compare Individual Gains
root-0-0-2-3-0-1-4-4: Formulate Collaboration Tactics
root-0-0-2-3-0-2: Detect Potential Competition
root-0-0-2-3-2: Strategic Adaptation
root-0-0-2-3-2-0: Analyze Other Players' Contributions
root-0-0-2-3-2-4: Flexibility in Decision Making
root-0-0-2-3-2-4-1: Adjust Contribution Based on Public Pot Size
root-0-0-2-3-2-4-2: Balance Risk and Reward
root-0-0-2-3-2-4-2-0: Assess the Current Token Balance
root-0-0-2-3-2-4-2-2: Adapt Contribution Strategy
root-0-0-2-3-2-4-2-4: Observe Patterns
root-0-0-2-3-3: Long-term Planning
root-0-0-2-3-4: Risk Assessment
root-0-0-2-3-4-0: Analyze Previous Rounds
root-0-0-2-3-4-0-1: Gain Assessment
root-0-0-2-3-4-0-2: Competitive Strategies
root-0-0-2-3-4-0-3: Collaboration Opportunities
root-0-0-2-3-4-2: Assess Potential Losses
root-0-0-2-3-4-4: Long-term Planning
root-0-0-2-4: Long-term Planning
root-0-0-2-4-0: Monitor Token Balance
root-0-0-2-4-0-0: Analyze Contribution Impact
root-0-0-2-4-0-0-2: Strategy Effectiveness
root-0-0-2-4-0-0-2-0: Contribution Analysis
root-0-0-2-4-0-0-2-0-2: Identify rounds with lower gain than expected and analyze potential reasons
root-0-0-2-4-0-0-2-0-3: Experiment with different contribution amounts in future rounds
root-0-0-2-4-4: Risk Management
root-0-0-2-4-4-0: Assess Potential Gains
root-0-0-2-4-4-0-0: Analyze Contribution Impact
root-0-0-2-4-4-1: Balance Contribution
root-0-0-2-4-4-3: Long-term Planning
root-0-0-2-4-4-4: Flexibility in Contributions
root-0-3: Adaptability
root-0-3-2: Observation and Prediction
root-0-3-2-1: Predict Potential Strategies
root-0-3-2-1-0: Player 1
root-0-3-2-1-1: Player 2
root-0-3-2-1-2: Player 3
root-0-3-2-2: Adjust Your Strategy
root-0-3-2-4: Stay Flexible
root-0-3-3: Risk Assessment
root-0-3-3-1: Consider Contribution Variability
root-0-3-3-1-1: Predict Potential Contributions
root-0-3-4: Long-term Adaptation
root-0-3-4-2: Flexibility in Contribution
root-0-3-4-2-2: Balance Short-term Gains and Long-term Goal
root-0-4: Risk Assessment
root-0-4-0: Analyze Previous Rounds
root-0-4-0-1: Risk Assessment
root-0-4-0-1-0: Analyze Previous Rounds
root-0-4-0-1-1: Consider Variability
root-0-4-0-1-3: Risk Tolerance
root-0-4-0-1-4: Strategic Adjustment
root-0-4-0-3: Strategic Planning
root-0-4-4: Adaptation
root-1: Strategic Decision Making
root-1-0: Analyze Other Players' Contributions
root-1-0-3: Consider Overall Game Dynamics
root-1-0-3-1: Assess Token Distribution
root-1-1: Consider Potential Payoff
root-1-1-2: Risk Assessment
root-1-1-2-0: Analyze Previous Rounds
root-1-1-2-0-0: Contribution Level Analysis
root-1-1-2-0-2: Trend Identification
root-1-1-2-0-2-0: Consider the overall game dynamics
root-1-1-2-0-2-1: Flexibility in contribution strategies
root-1-1-2-0-2-2: Risk management
root-1-1-2-0-2-2-0: Analyze Trends
root-1-1-2-0-2-2-2: Diversify Contributions
root-1-1-2-0-2-3: Observation of player behavior
root-1-1-2-0-3: Risk Assessment
root-1-1-2-0-4: Adaptation Strategy
root-1-1-2-0-4-2: Consider Overall Game Dynamics
root-1-1-2-4: Long-term Risk Management
root-1-1-3: Adapt to Player Behaviors
root-1-1-3-2: Strategic Decision Making
root-1-3: Adapt to Player Behaviors
root-1-3-3: Balance Risk and Reward
root-1-5: Flexibility
root-1-5-1: Adjust Contribution Based on Public Pot
root-1-5-1-0: Analyze Public Pot Size
root-1-5-1-0-2: Monitor Overall Trends
root-1-5-1-0-2-2: Compare with Other Players
root-1-5-1-2: Monitor Overall Token Accumulation
root-2: Long-term Planning
root-2-0: Assess Previous Contributions
root-2-0-1: Identify Optimal Contribution Levels
root-2-0-2: Consider Player Behaviors
root-2-0-3: Adjust Contribution Strategy
root-2-1: Strategic Contribution
root-2-2: Monitor Other Players


Guess 2/3 of the Average:

root: Choose a number that you believe will be closest to 2/3 of the average of all numbers chosen by players, including your selection
root-0: Observation
root-0-0: Analyze Trends
root-0-0-1: Evaluate Deviations
root-0-0-1-3: Stay Informed
root-0-0-1-3-3: Flexibility in Decision-Making
root-0-0-1-3-3-1: Adapt to Changing Dynamics
root-0-0-1-3-3-1-3: Consider Risk-Reward
root-0-0-1-3-3-2: Consider Risk-Reward Tradeoff
root-0-0-1-3-3-2-3: Adapt to Changing Circumstances
root-0-0-1-3-3-2-3-3: Strategic Observation
root-0-0-1-3-3-2-3-3-1: Consider Recent Rounds
root-0-0-1-3-3-2-3-3-2: Identify Outliers
root-0-0-1-3-3-2-3-3-3: Predict Potential Average
root-0-0-1-3-3-2-3-4: Risk Assessment
root-0-0-1-3-3-4: Balance Consistency and Adaptability
root-0-0-1-3-4: Strategic Observation
root-0-0-1-3-4-0: Analyze Winning Numbers
root-0-0-1-3-4-0-1: Identify Common Numbers
root-0-0-1-3-4-0-2: Consider the Average
root-0-0-1-3-4-1: Monitor Average Numbers
root-0-0-1-3-4-1-2: Consider Previous Results
root-0-0-1-3-4-1-4: Adjust Risk Tolerance
root-0-0-1-3-4-2: Observe Your Performance
root-0-0-1-3-4-3: Consider Player Strategies
root-0-0-1-3-4-3-0: Analyze Winning Strategies
root-0-0-1-3-4-3-1: Adaptation
root-0-0-1-3-4-3-2: Observation
root-0-0-1-3-4-3-4: Risk Assessment
root-0-1: Identify Outliers
root-0-1-0: Analyze Previous Rounds
root-0-1-0-1: Consider Trends
root-0-1-0-1-0: Consider the decreasing trend in the average number chosen by players in the previous rounds and select a number slightly lower than the expected average for the upcoming round
root-0-1-0-1-0-3: Balance Risk and Reward
root-0-1-0-1-0-3-2: Cautious Approach
root-0-1-0-1-0-3-3: Strategic Thinking
root-0-1-0-1-0-3-5: Observation
root-0-1-0-1-0-4: Monitor Results
root-0-1-0-2: Adjust for Variability
root-0-1-0-2-0: Analyze Previous Averages
root-0-1-0-2-0-1: Identify Trends
root-0-1-0-2-0-1-2: Consider the Range
root-0-1-0-2-0-2: Consider Outliers
root-0-1-0-2-0-2-0: Analyze Previous Outliers
root-0-1-0-2-0-2-3: Factor in Player Behavior
root-0-1-0-2-0-2-3-1: Identify Player Tendencies
root-0-1-0-2-0-2-3-2: Adjust Number Selection
root-0-1-0-2-1: Consider Conservative Approach
root-0-1-0-2-1-1: Identify Central Tendency
root-0-1-0-2-1-2: Avoid Extreme Outliers
root-0-1-0-2-1-3: Consider Stability
root-0-1-0-2-1-4: Balance Risk and Reward
root-0-1-0-2-1-4-1: Consider the Current Average
root-0-1-0-2-1-4-2: Assess Your Position
root-0-1-0-2-1-4-4: Adapt to the Game Dynamics
root-0-1-0-2-1-4-5: Stay Informed
root-0-1-0-2-2: Evaluate Trends
root-0-1-0-2-4: Adapt to Changing Dynamics
root-0-1-0-2-4-1: Flexibility in Number Selection
root-0-1-0-2-4-2: Consider Outliers
root-0-1-0-2-4-4: Risk Assessment
root-0-1-1: Consider Potential Influences
root-0-1-2: Predict Potential Outliers
root-0-1-2-0: Analyze the Trend
root-0-1-3: Adjust Your Strategy
root-0-1-3-1: Consider the Trend
root-0-1-3-1-1: Adjust Strategy
root-0-1-3-1-2: Stay Vigilant
root-0-1-3-2: Balance Risk and Reward
root-0-1-3-2-1: Consider the Impact of Outliers
root-0-1-3-2-1-0: Analyze Previous Rounds
root-0-1-3-2-1-1: Adjust Strategy
root-0-1-3-2-1-2: Monitor Extreme Numbers
root-0-1-3-2-1-4: Stay Flexible
root-0-1-3-2-4: Stay Informed
root-0-1-3-3: Adapt to Competitors
root-0-1-3-3-1: Balance Risk and Reward
root-0-1-3-3-2: Anticipate Competitors' Choices
root-0-1-3-3-2-4: Flexibility
root-0-1-3-3-4: Strategic Risk-Taking
root-0-1-3-3-4-2: Consider the Range
root-0-1-3-3-4-3: Balance Consistency and Differentiation
root-0-1-3-3-4-4: Adapt Based on Previous Outcomes
root-0-2: Consider Player Behavior
root-0-2-1: Adjust Based on Averages
root-0-2-3: Stay Flexible
root-0-2-3-2: Evaluate Your Position
root-0-2-3-3: Monitor Player Behaviors
root-0-3: Factor in Previous Results
root-0-3-1: Consider Trend
root-0-4: Adjust Strategy
root-0-4-1: Consider Your Competitors
root-0-4-1-1: Adjust for Biases
root-0-4-1-3: Use Game Theory
root-0-4-1-3-1: Anticipate Competitors' Choices
root-0-4-1-3-3: Consider Risk-Reward
root-0-4-3: Stay Informed
root-0-4-4: Utilize Strategic Thinking
root-1: Strategic Thinking
root-1-2: Calculating 2/3 of the Average
root-1-3: Strategic Number Selection
root-1-4: Adaptation and Flexibility
root-1-4-2: Evaluate Your Own Strategy
root-1-4-4: Stay Informed
root-1-4-5: Strategic Variation
root-2: Risk Assessment
root-2-1: Consider Variability
root-2-3: Assess Risk Tolerance
root-2-4: Anticipate Strategic Play
root-3: Adaptation
root-3-3: Risk Assessment
root-3-3-1: Consider the Range
root-3-3-4: Utilize Previous Experience
root-4: Long-term Planning
root-4-2: Strategic Adjustment
root-4-4: Risk Assessment
root-4-4-1: Consider Variability
root-4-4-2: Evaluate Your Performance


Auction Arena:

root: secure the highest profit at the end of this auction, compared to all other bidders
root-0: Efficiently allocate budget
root-0-0: Prioritize items with a higher difference between your estimated value and the starting price
root-0-0-1: Consider the competition
root-0-0-1-1: Identify Weaknesses
root-0-0-1-1-1: Monitor Budget Utilization
root-0-0-1-1-1-1: Strategically Allocate Bids
root-0-0-1-1-1-1-2: Monitor Competitor Bids
root-0-0-1-1-1-1-2-1: Strategic Allocation of Bids
root-0-0-1-1-1-1-2-1-1: Focus on Items with Less Interest
root-0-0-1-1-1-1-2-1-2: Monitor Potential Withdrawals
root-0-0-1-1-1-1-2-2: Budget Conservation
root-0-0-1-1-1-4: Maintain Flexibility
root-0-0-1-1-2: Assess Risk-Taking Behavior
root-0-0-1-1-2-1: Identify Weaknesses
root-0-0-1-1-2-1-0: Analyze Bidding Patterns
root-0-0-1-1-2-1-3: Monitor Remaining Items
root-0-0-1-1-2-3: Budget Management
root-0-0-1-1-3: Identify Overestimation
root-0-0-1-1-4: Exploit Predictable Behavior
root-0-0-1-2: Formulate Counter-Strategies
root-0-0-1-2-4: Psychological Tactics
root-0-0-1-3: Adaptability
root-0-0-1-3-1: Adjust Bidding Strategy
root-0-0-1-3-4: Evaluate Risk-Reward Ratio
root-0-0-1-5: Information Utilization
root-0-0-1-5-0: Analyze Bidders' Behavior
root-0-0-1-5-1: Adjust Bidding Strategy
root-0-0-1-5-1-0: Analyze Previous Bidding Patterns
root-0-0-1-5-1-0-1: Target Items with Lower Competition
root-0-0-1-5-1-0-3: Evaluate True Values
root-0-0-1-5-1-2: Evaluate Profit Margins
root-0-0-1-5-1-3: Identify High-Value Items
root-0-0-1-5-1-6: Adapt to True Values
root-0-1: Monitor the bidding behavior of other bidders
root-0-1-2: Strategic Bidding
root-0-1-2-5: Stay Informed
root-0-3: Be prepared to adjust your estimated value
root-0-4: Aim for a balance between winning bids and maximizing profit
root-1: Accurately estimate item values
root-1-0: Research
root-1-1: Analyze Previous Auctions
root-1-1-1: Analyze Market Trends
root-1-1-1-0: Research Market Demand
root-1-1-1-1: Consider Seasonality
root-1-1-1-2: Economic Conditions
root-1-1-2: Adjust Estimated Values
root-1-2: Consider Item Condition
root-1-3: Adjust Estimations
root-1-3-1: Consider True Value
root-1-3-4: Adapt to Competition
root-1-4: Budget Management
root-1-4-1: Risk Assessment
root-1-4-2: Prioritize High-Value Items
root-1-4-2-0: Assess Remaining Budget
root-1-4-2-3: Monitor Competing Bidders
root-1-5: Risk Assessment
root-2: Strategic bidding
root-2-0: Budget Management
root-2-1: Estimated Value Comparison
root-2-2: Observation of Competitors
root-2-3: Risk Assessment
root-2-4: Strategic Withdrawal
root-2-4-0: Assess Potential Profit Margin
root-2-4-5: Long-term Profit Maximization
root-3: Risk management
root-3-1: Budget Allocation
root-3-2: Competitive Analysis
root-3-2-1: Assess Remaining Competitors
root-3-2-2: Estimate Competitors' Valuation
root-3-3: Flexibility in Bidding
root-3-5: Information Gathering
root-3-5-1: Refine risk assessment
root-3-5-4: Anticipate competition
root-3-5-5: Adapt bidding strategy
root-4: Adaptability
root-4-4: Risk Management
root-4-6: Adapt to Market Dynamics

DealOrNotDeal

root: minimize the profit gap between yourself and your partner in this negotiation, regardless of your own profit.
root-0: Maximize the number of items you receive
root-0-0: Evaluate the value of each item
root-0-1: Consider trade-offs
root-0-2: Seek compromise
root-0-3: Communicate effectively
root-0-4: Be flexible
root-1: Prioritize high-value items
root-1-0: Assess the value of each item
root-1-1: Consider trade-offs
root-1-2: Negotiate for high-value items
root-1-3: Be open to compromise
root-1-4: Communicate the reasoning behind your prioritization
root-2: Ensure fair distribution
root-2-0: Consider the value of each item
root-2-1: Propose a balanced allocation
root-2-2: Be open to compromise
root-2-3: Communicate the reasoning behind your proposal
root-2-4: Seek mutual agreement
root-3: Maintain a cooperative and communicative approach
root-3-0: Clarify interests and priorities
root-3-1: Seek common ground
root-3-2: Explore trade-offs
root-3-3: Remain open to creative solutions
root-3-4: Maintain a positive and respectful tone
root-4: Adapt and adjust strategies
root-4-0: Understand Bob's priorities
root-4-2: Propose alternative allocations
root-4-3: Maintain open communication
root-4-4: Be willing to compromise

\end{lstlisting}

\label{sec:appendix}

\end{document}